\documentclass[runningheads]{llncs}
\usepackage{graphicx}
\usepackage{float}
\usepackage{amsmath}

\usepackage{algorithm}
\usepackage{algpseudocode}
\usepackage[section]{placeins}

\usepackage{etoolbox}
\usepackage{ifthen}
\usepackage[dvipsnames]{xcolor}

\usepackage{hyperref}

\newboolean{showcomments}
\setboolean{showcomments}{false}

\ifthenelse{\boolean{showcomments}}
  {\newcommand{\nb}[3]{
  {\color{#2}#1}
  {\color{#2}{#3}}
  }
  }
  {\newcommand{\nb}[3]{}
  }

\newcommand\konstantin[1]{\nb{\textbf{Konstantin:}}{blue}{#1}}

\begin{document}
\title{Planning to Score a Goal in Robotic Football with Heuristic Search \thanks{ Camera-ready version of the paper as to appear in ICR 2020 proceedings}}
\titlerunning{Planning to Score a Goal in Robotic Football with Heuristic Search}
%
\author{Ivan Khokhlov\inst{1} \and
Vladimir Litvinenko\inst{1} \and
Ilya Ryakin\inst{1} \and
Konstantin Yakovlev\inst{1, 2}
}
\authorrunning{I. Khokhlov et al.}
%
\institute{Moscow Institute of Physics and Technology, Dolgoprudny, Russia
\and
Federal Research Center for Computer Science and Control of Russian Academy of Sciences, Moscow, Russia
\\
\email{khokhlov.iyu@gmail.com, litvinenko.vv@phystech.edu, ryakin.is@phystech.edu, yakovlev@isa.ru}
}
%
\maketitle              
\begin{abstract}
This paper considers a problem of planning an attack in robotic football (RoboCup). The problem is reduced to finding a trajectory of the ball from its current position to the opponents goals. Heuristic search algorithm, i.e. A*, is used to find such a trajectory. For this algorithm to be applicable we introduce a discretized model of the environment, i.e. a graph, as well as the core search components: cost function and heuristic function. Both are designed to take into account all the available information of the game state. We extensively evaluate the suggested approach in simulation comparing it to a range of baselines. The result of the conducted evaluation clearly shows the benefit of utilizing heuristic search within the RoboCup context.
\end{abstract}
\section{Introduction}

Robotic football competitions has been one of the prominent drivers of the robotic research since 1997. Teams of robots that play football against each other face a wide range of challenging problems: locomotion, path and motion planning, communication, localization, interaction, and many others. The idea of organizing a competition between robots playing football emerged in early 90s of XX century and since them transformed to a global initiative called \emph{Robocup} with regular tournaments, different leagues and more than 3500 participants representing major universities, research institutes and commercial organizations involved in robotic research\footnote{For a brief history of Robocup initiative refer to \url{https://www.robocup.org/a\_brief\_history\_of\_robocup}.}. All RoboCup community is united by a big goal -- in 2050 the champion of RoboCup Humanoid football competition should be able to play against human champions of FIFA World Cup according to FIFA rules.

Nowadays, RoboCup Humanoid football rules is quite simpler than FIFA one. Field of the is $6\times9$ meters, covered by 30 mm height grass and marked up with white lines.  Robots must be similar to human in sensors, body structure, proportions and even center of mass position. Teams of 4 robots compete on the field each for two 10 minute halves. Moreover, they can communicate via Wi-Fi network with each other and referee.
\konstantin{I don't think that the paragraph above is necessary in Intro section. May be will use it further in text.}

\begin{figure}[t]
    \centering
    \includegraphics[width=\linewidth]{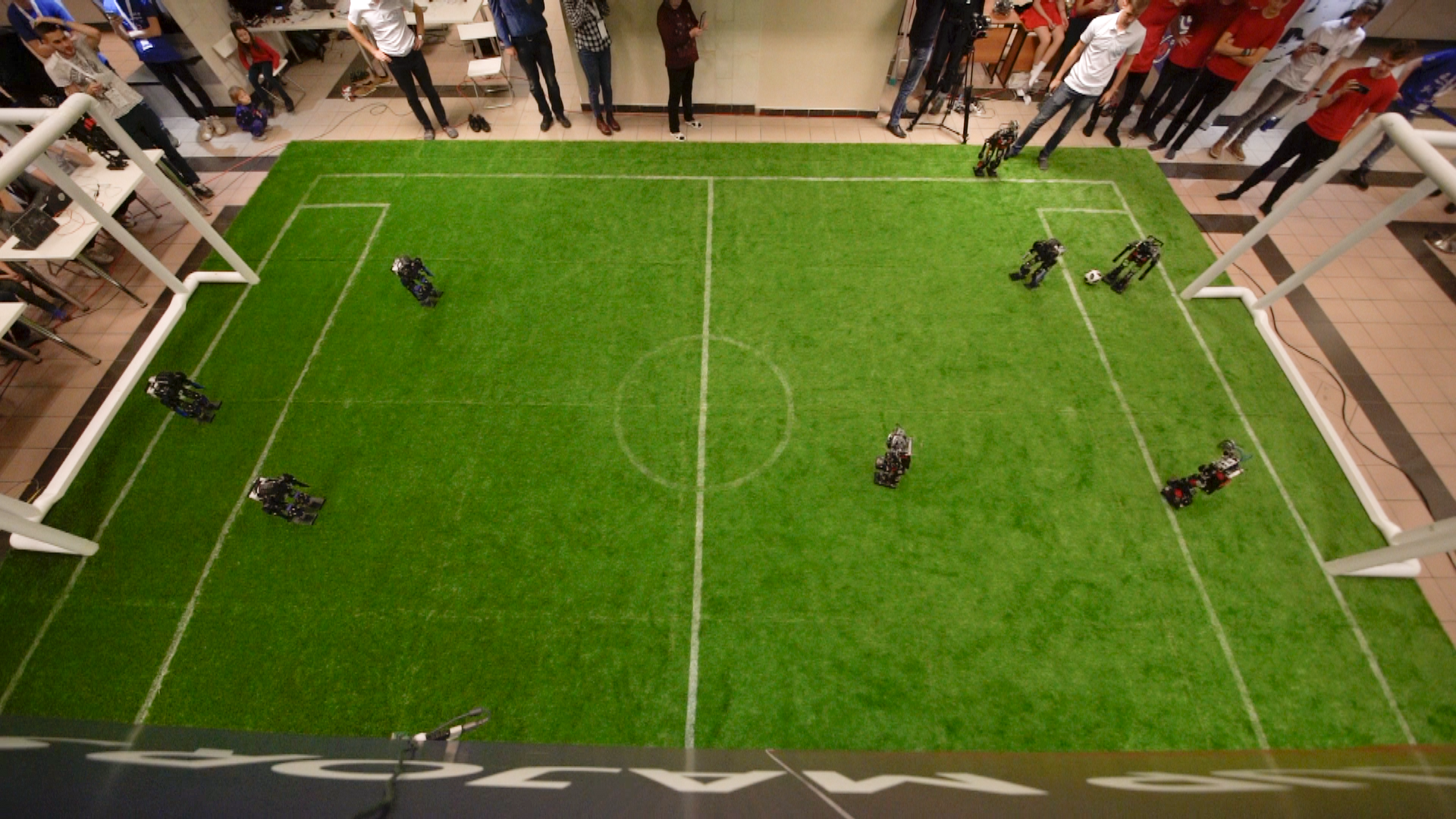}
    \label{fig:workplace}
    \caption{Robotic football setup: humanoid-robots playing the ball on the reduced copy of a footbal field.}
\end{figure}

Intrinsically, robotic humanoid football is very challenging domain. The problems that arise here can be roughly decomposed into two-level hierarchy. First, one needs to ensure stable locomotion, consistent detection and localization of the key object of interest, i.e. the ball, the posts, the opponents etc., reliable peer-to-peer communication between the robots.
Second, the more involved problems such as role assignment and planning to score a goal arise. In this work we are interested mainly in the latter problems and consider the first ones to be successfully solved with a desired degree of accuracy \cite{tdp}. 

Specifically we are focused on the kick planning for attack phase of the game. This problem constantly arise within the game when our team intercepts a ball and aims to score a goal which is vital for winning. We approach this problem by boiling it down to path planning for a ball. That is, we suggest to use a graph-based heuristic search algorithm to find a shortest path for a ball from its current location to the opponent's goals. We evaluate this approach empirically in simulation and compare it to a range of the baseline strategies. We show that the suggested approach outperforms the competitors in a wide variety of different game scenarios.

\section{Related Work}

Initially much of the research in robotic football was concentrated around locomotion, tracking, localization etc. More recently, the teams competing in RoboCup Humanoid League have started to put more emphasis on the tactics and strategy. Meanwhile, in 2D and 3D Simulation Leagues solving them for more than 10 years. For example, noteworthy approaches with using predefined game strategies were introduced in \cite{predefined1} and \cite{predefined2}. The authors figured out that template attacks increase game quality and developed framework for fast programming such strategies. In \cite{forwardSim} the authors consider problem similar to ours. They assume kick target to be modelled as Gaussian distribution. With such probabilistic comprehension authors calculate probability of the ball to be in one of 6 states and for each state calculated weight in predefined potential field. 

Works that consider the application of heuristic search to path and motion planning for humanoid robots, not necessarily within the robotic football context, are more numerous. For example, \cite{gutmann2005real} describe application of the A* algorithm to planning possible paths through uneven terrain. In \cite{footstepplanning}, \cite{HOAP} and \cite{hornung2013search} a more involved problem of planning steps of the humanoid robot is considered. \cite{wermelinger2016navigation},  \cite{ADFA} and \cite{Rerouting} study the combined problem of global path planning and footstep planning for legged robots. A comparison of different heuristic search algorithms applied to both of the aforementioned problems can be found in \cite{arain2013comparison}.

\section{Problem Statement}
Consider two teams of humanoid robots playing football on a plain field sized $6 \times 9$ meters and covered with 30 mm artificial grass. Each team is composed of the 4 robots with body mass index in the range $[3, 30]$ and height under 1 meter\footnote{Such robots are attributed as \emph{kid-sized} in Robocup competition.}. 
The ultimate aim of a team is to win the game which is achieved via scoring more goals than the opponent. The goal posts are located on the opposite sides of the field and are 2.6 meters in width, the ball is 20 cm in diameter. The typical setup is depicted in \figurename~\ref{fig:workplace}.

Game controller that manipulates the robots of our team constantly localizes them, as well as the ball and the opponents (so we consider all these positions to be known). Assume now that the ball is close to one our robots (i.e. the distance between the ball and one of our robots is much shorter than the one between the ball and any of the opponents) and we want to start an attack. The latter is understood as a sequence of kicks made by our robots with the aim of scoring a goal. The problem now is to plan an attack and, more specifically, to estimate the direction of a first kick in such a way that \emph{i}) the kick won't result in loosing a ball (i.e. it will be our robot that will be the first on the ball, not the opponent), \emph{ii}) the kick results in a ``winning position''. The later intuitively means that the chances of scoring a goal after the kick increase.
\konstantin{Don't like the last phrases. Look too informal. Need to think about it.}

In this work we reduce the problem of attack planning to \emph{finding a path for a ball} from its current position to the ``in-the-net'' position. This path is a sequence of segments. Each segment represents a  ball trajectory after a kick of a predefined force performed by one of the robots of our team. The first segment of the path should not intersect the areas occupied by opponent's robots. The last segment should intersect the boundary of the field in between the opponent's poles. The criterion to be minimized is time to the goal.

\section{Method}

We rely on a graph search algorithm, i.e. A*~\cite{hart1968formal}, to solve the considered path finding problem. For this algorithm to be applied we need to \emph{i}) define a graph; \emph{ii}) define such search components as a cost function and a heuristic function to be used within the search. We describe these components next.

\subsection{Graph}

\begin{figure}[t]
    \centering
    \includegraphics[width=0.5\textwidth, angle=90]{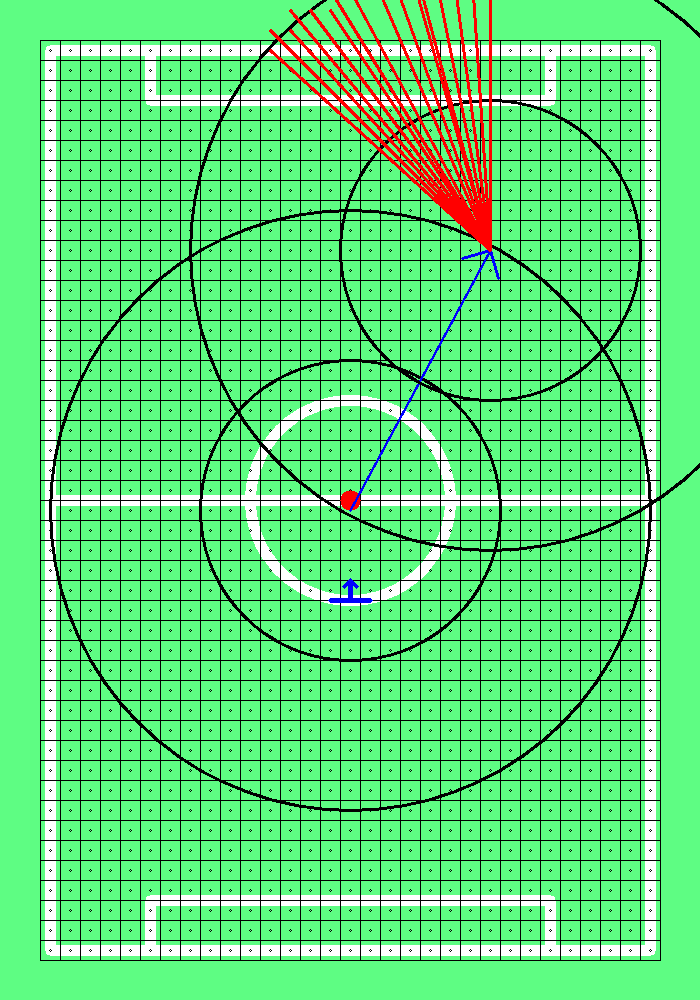}
    \label{fig:field}
    \caption{Graph used for path planning. Centers of the grid cells define the vertices. Edges are defined implicitly by each pair of the vertices that (approximately) lie within the predefined kick distance from each other.}
    
\end{figure}

We introduce a graph by discretizing the workspace, which is a 2D rectangle sized $6 \times 9$ meters, via the cell decomposition. Each cell is a square of $10 \times 10$ cm, so there is $90 \times 60 = 5 400$ cells overall. The center of each cell defines a graph vertex. The current position of a ball, given to a path planner by the external localization system as a tuple $(x, y)$, is tied to a graph vertex in the following fashion. Knowing $(x, y)$ we identify the cell which center is the closest to the ball's position and then assume the ball to be located in the center of this cell. If there are different cells which centers are equidistantly close to the ball we choose one of these cells, as the start vertex, arbitrarily.

The edges of a graph are defined as follows. We assume that a robot can kick the ball with a predefined force, so the distance travelled by the ball, $r_{kick}$, is proportional to that force. When a ball is at a graph vertex $v$ we identify all vertices $v'$ that form a discrete approximation of a circumference of radius $r_{kick}$ -- see~\figurename~\ref{fig:field}. Each tuple $(v, v')$ defines a graph edge now. Moreover if a kick ends beyond the field but the ball travels in between the opponent's goal posts the corespondent edge is also considered to be part of the graph (a few examples of such edges are shown in red in~\figurename~\ref{fig:field}).

Indeed, the overall number of edges in the introduced graph depends on the value of $r_{kick}$ and can be very high. Moreover, different values for $r_{kick}$ can be allowed, which contributes to increasing the number of edges. Thus we do not store it explicitly but rather implicitly construct the edges while the search.

\subsection{Search}

The input of the search algorithm is a graph (as defined above) as well as the positions of our robots and the robots of the opponent. The output is expected to be a least cost path in that graph that starts in the vertex associated with the current location of the ball and ends with an edge that lies in between the goal posts of the opponent. The cost of the path is the cumulative cost of the edges forming that path, thus we need to define how the cost of an individual edge is computed.

\subsubsection{Cost of an edge}

\begin{algorithm}[t]
\caption{Cost function}\label{alg:costFunc}
\begin{algorithmic}

\Function{computeCost}{robotPos, opponentsPositions, ballFromPos, ballToPos, firstKick}
\State $ballTravelTime\gets getLength(toPos, fromPos)/ballSpeed$
\If {firstKick}
    \State $timeToReachBall \gets calcTimeToApproachBall(fromPos, robotPos)$
    \If {intersectOpponent(fromPos, toPos, opponentsField)}
        \State \Return $timeToReachBall + ballTravelTime * 2$
    \Else
        \State \Return $timeToReachBall + ballTravelTime $
    \EndIf
\Else
    \State \Return $ballTravelTime$
\EndIf
\EndFunction
\end{algorithmic}
\end{algorithm}

Recall, that each graph edge represents a kick performed by a robot, thus a cost of an edge is associated with the time needed for this kick to be accomplished, i.e. the time by which the ball reaches the endpoint of a kick. 

In most cases we compute an edge's cost by dividing its length to the speed of the ball\footnote{We assume a simplistic ball movement model when the ball moves with a constant speed}. However, computing cost of the edges that have the start vertex as their endpoint is more involved. The rationale behind this is that we know the positions of the opponent's robots as well as the location of our robots when we start at attack, thus, it's reasonable to take this information into account.

When the first kick is performed we add to the correspondent edge cost the time that the kicking robot of our team (the one that is closest to the ball) will spend on approaching the ball. Moreover we penalize kicks that have a high risk of being intercepted by the opponent. Recall, that all positions of the opponent's robots are known. We model these robots as disks and compute whether an edge, representing the first kick, intersects any of them. If this is the case the cost of the edge is multiplied by a constant factor (we use 2 in our experiments). Thus the resultant path is less likely to contain such an edge. The reason we do not prune such edges for good is that the positions of the robots, reported by the external tracking system, are not 100\% accurate in practice. 

The high-level algorithm that computes the cost of an edge, associated with a kick, is presented in Alg.~\ref{alg:costFunc}.

\subsubsection{Heuristic function}

\begin{algorithm}[t]
\caption{Heuristic function}\label{alg:hFunc}
\begin{algorithmic}

\Function{hFunc}{teamMatesField, toPos, firstKick}
\State $timeToReachGoal\gets distToGoal(toPos)/ballSpeed$
\If {firstKick}
    \State $timeToApproachBall \gets calcTimeToApproachBall(toPos, teamMatesField)$
    \State \Return $timeToApproachBall + timeToReachGoal$
\Else
    \State \Return $timeToReachGoal$
\EndIf
\EndFunction
\end{algorithmic}
\end{algorithm}

Heuristic function takes as input the position of the ball, i.e. the graph vertex, and outputs the lower bound of time needed for the ball to reach the opponent's gates.

Similarly to the cost function, we compute such a heuristic estimate in most cases in a straightforward fashion. First, we calculate the distance form the ball position (vertex in the graph) to the gates by using the closed-loop formula for computing the distance between the point (ball's position) and the line segment (opponents gates). Second, we divide this distance by the ball speed.

As before, we also introduce a more involved procedure for computing heuristic for a first kick. After that kick is made it will take some time by our next kicking robot to approach the ball to continue an attack, so it's reasonable to incorporate this information into the search process and make the heuristic function more informative. To do so we identify the robot of our team that is the closest to the endpoint of an edge that represent the first kick, compute the time needed for that robot to approach the ball to perform the next (second) kick, add this time to the heuristic estimate.

The high-level algorithm that computes the cost of an edge, associated with a kick, is presented in Alg.~\ref{alg:hFunc}.


\subsubsection{Heuristic search} Having defined the cost function and the heuristic function, we employ the renowned \textsc{A*} algorithm to compute the least cost path in the given graph. This path corresponds to the minimal-time trajectory of the ball from its current position to the gates of the opponent. Please note, that we do not simulate the moves of our robots and the moves of the opponent when finding such a trajectory, thus it is likely to become inaccurate after the attack evolves. At the same time, the computational budget needed to accomplish the suggested search is very low, thus one can invoke re-planning after each kick to keep the plan updated.


\section{Empirical Evaluation}

We ran empirical evaluation of the suggested approach in a simulation of a football game, i.e. we placed our robots, enemies and the ball on the field and started the game from this layout. During the simulation we assumed perfect execution, i.e. all commands sent to robot actuators were executed perfectly. We also assume perfect localization, i.e. each robot localized itself, allies enemies and the ball perfectly. All kicks were considered to be performed precisely as well. In our evaluation we separately run test for 4m kicks and 2m kicks. These values were chosen based on our experience with kick controllers of real robots.

We compared four different strategies to estimating the direction of a kick during the attack:

\begin{itemize}
    \item \textit{Planning} Our strategy as described above. The robot closest to the ball performs the heuristic search and as a result gets the sequence of kicks from the ball's current position to the ``in-the-net'' position. Then the first kick from that sequence is made.
    \item \textit{Reactive} This strategy chooses the most promising kick based on the one-step look-ahead planning. It can be seen as a ``capped planning'' when only the first kick is planned not the whole attack.
    \item \textit{Forward} According to this strategy the kick is always made towards the enemy goals without taking into account any data regarding the positions of the enemies and/or allies.
    \item \textit{Expert} This approach was developed by \emph{Rhoban Team} \cite{allali2019rhoban}. The field is divided into the blocks sized $20~cm \times 20~cm$ and for each block, the direction of possible kicks are specified and ranged according to the expert score. The kick with the highest score that does not intersect an opponent is chosen for an attack. 
\end{itemize}

Each football game was simulated until the goal was scored by our team (\textit{success}) or until the ball was intercepted by the opponents (\textit{failure}). The interception may have occurred in two ways. First, an opponent might be the first to approach the ball after the kick had been made by our robot. Second, our robot might perform a kick that intersects an opponents zone, i.e. a circumference of a certain radius (we set it to be $20~cm$ in our experiments) centered at the position of an enemy robot (which is known from the perfect localization as described above). In real game, however, when the positions of the opponent robots are not precisely known, such a kick may not actually lead to an interception. To compensate for this and make the simulation more close to reality we tweaked the simulation as follows. Each time the ball passed trough an interception zone associated with some of the enemy robots we let it go through with the $0.5$ probability.

\begin{figure}[t]
    \centering
    \includegraphics[width=0.49\textwidth]{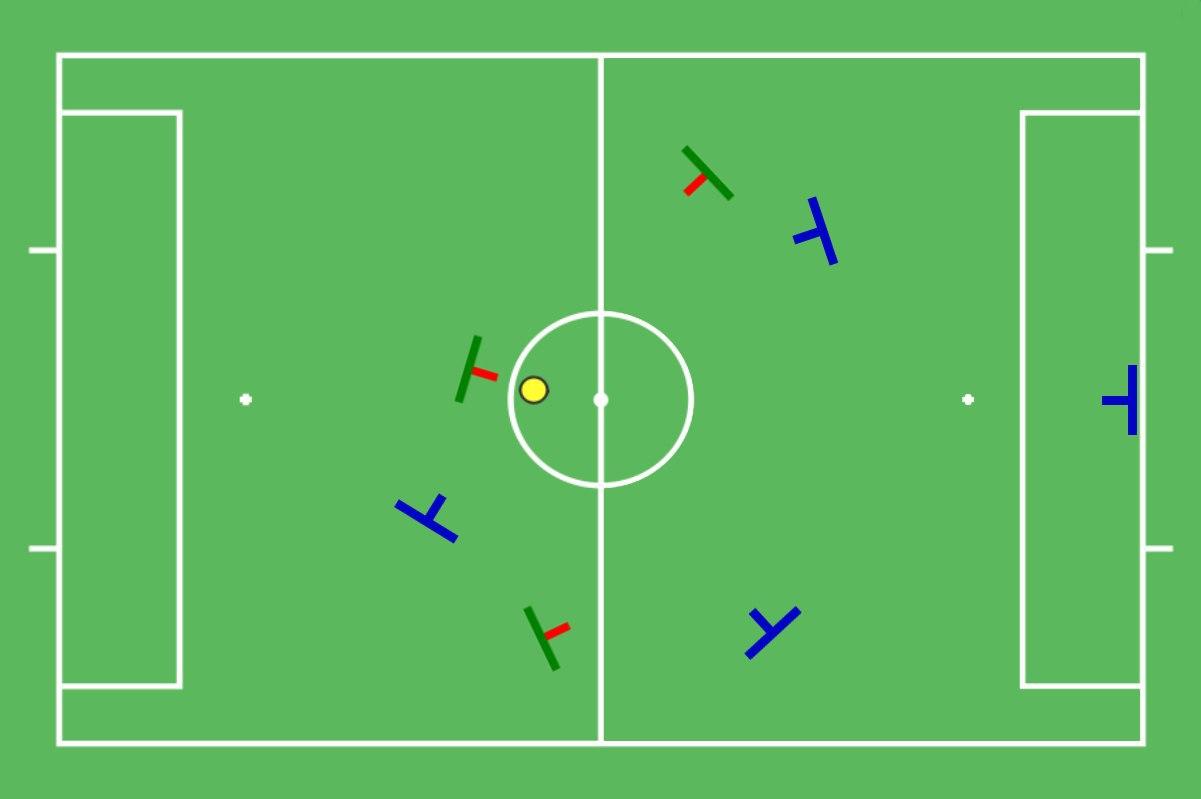}
    \includegraphics[width=0.49\textwidth]{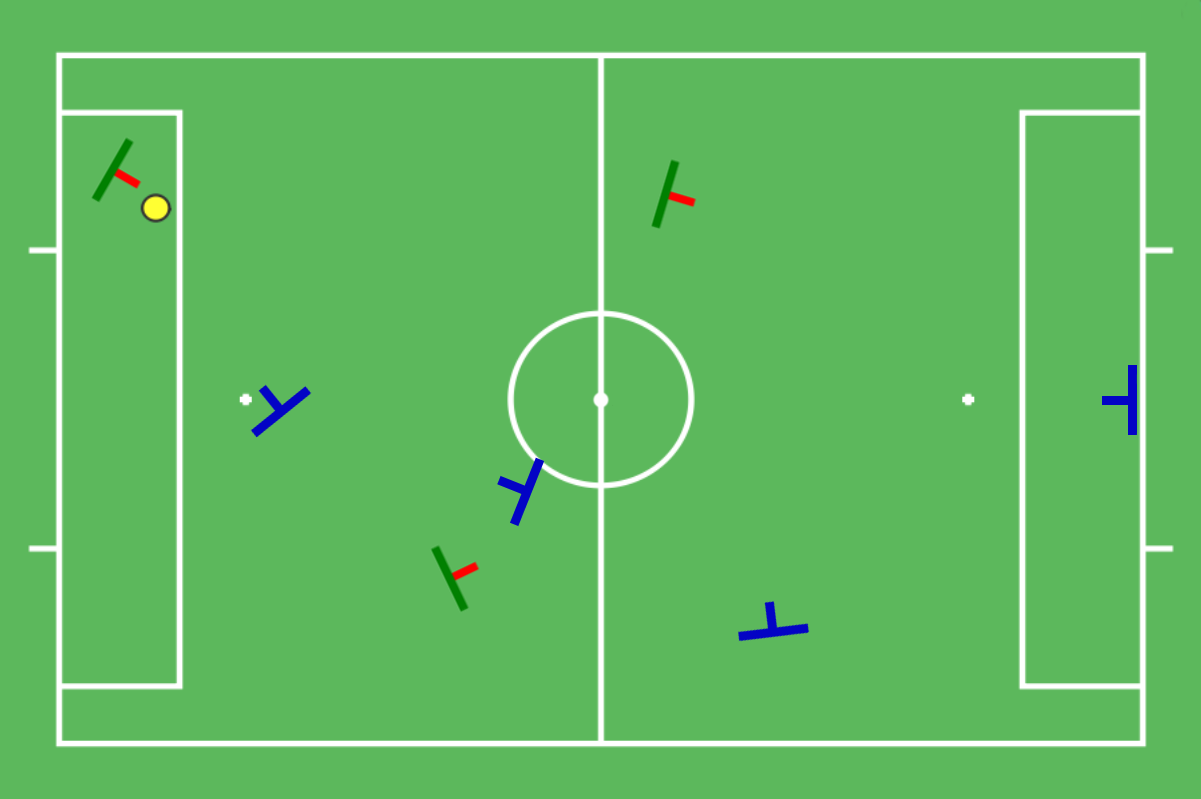}
    \caption{Different game layouts used in the experimets: random (left), attack (right)}
    \label{fig:layouts}
\end{figure}

We used two different types of game layouts for the experiments: \textit{random} and \textit{attack}. For random layout we placed 3 field robots of our team and 3 field robots of the opponent's team randomly. We also placed an enemy goal keeper appropriately (in the middle of the goals). The ball was placed at the distance of $20~cm$ from one of our robots. An example of this layout is shown on~\figurename~\ref{fig:layouts} (left).

Attack layout represents a typical phase of the game when an enemy attack has just finished and we got control over the ball and want to start an attack. One robot of our team is placed nearby our penalty zone and looks towards the enemy's goals. The ball is nearby this robot. Two other robots of our team are placed in the middle of the field, waiting for the pass. Speaking of the enemy team, we put one of its robots to our penalty zone, one to the middle of the field, and one robot was placed randomly. The enemy's goalkeeper was placed in the middle of the goals. An example of this layout is depicted on~\figurename~\ref{fig:layouts} (right).

Overall, we generated 100 different layouts of each type. Thus 200 football games in total were simulated.

\subsection{Results}

\begin{figure}[t]
\begin{minipage}{0.49\textwidth}
    \includegraphics[width=\textwidth]{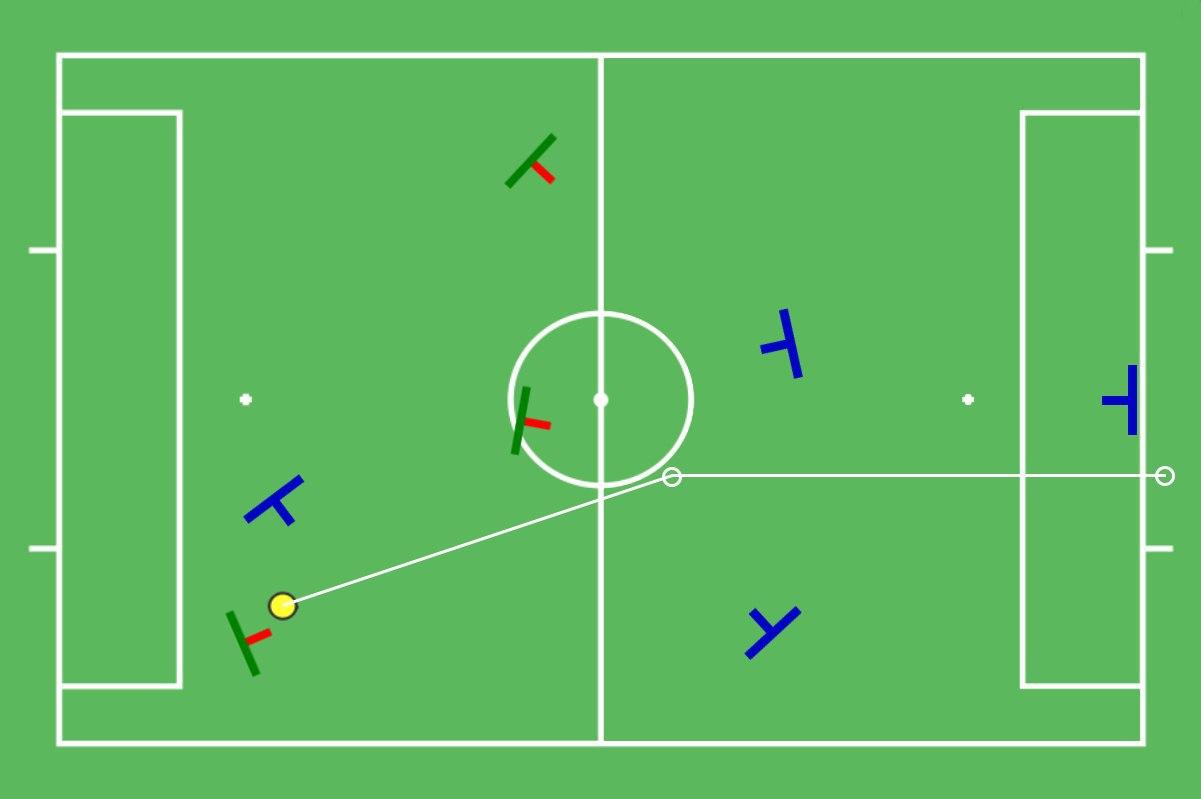}
    \includegraphics[width=\textwidth]{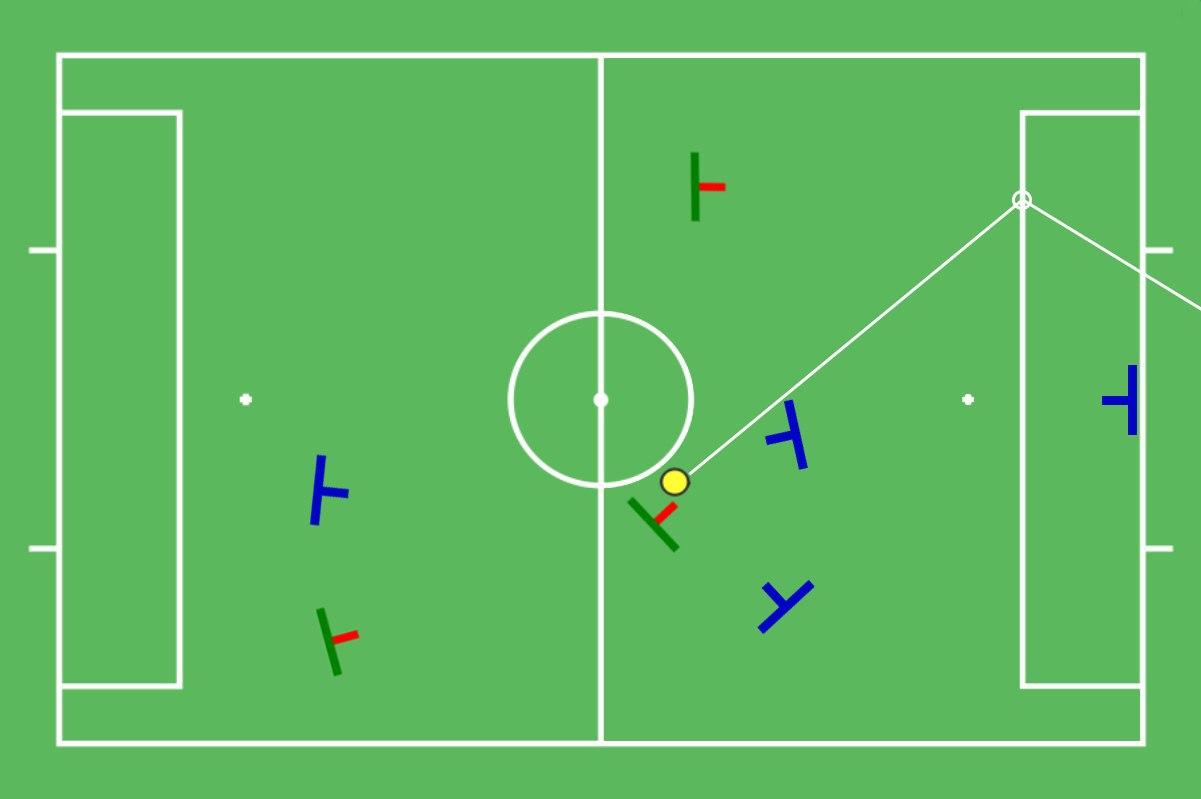}
    \includegraphics[width=\textwidth]{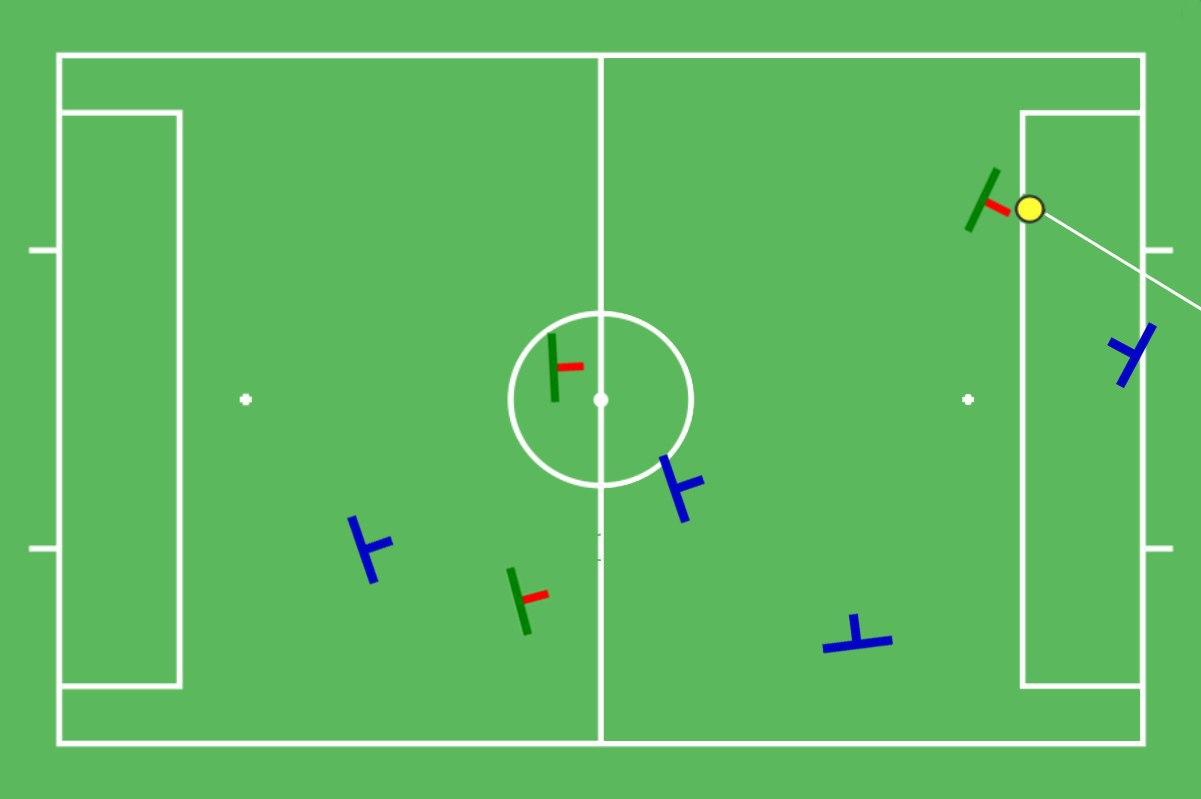}
    \label{fig:planning}
\end{minipage}
\hfill
\begin{minipage}{0.49\textwidth}
    \includegraphics[width=\textwidth]{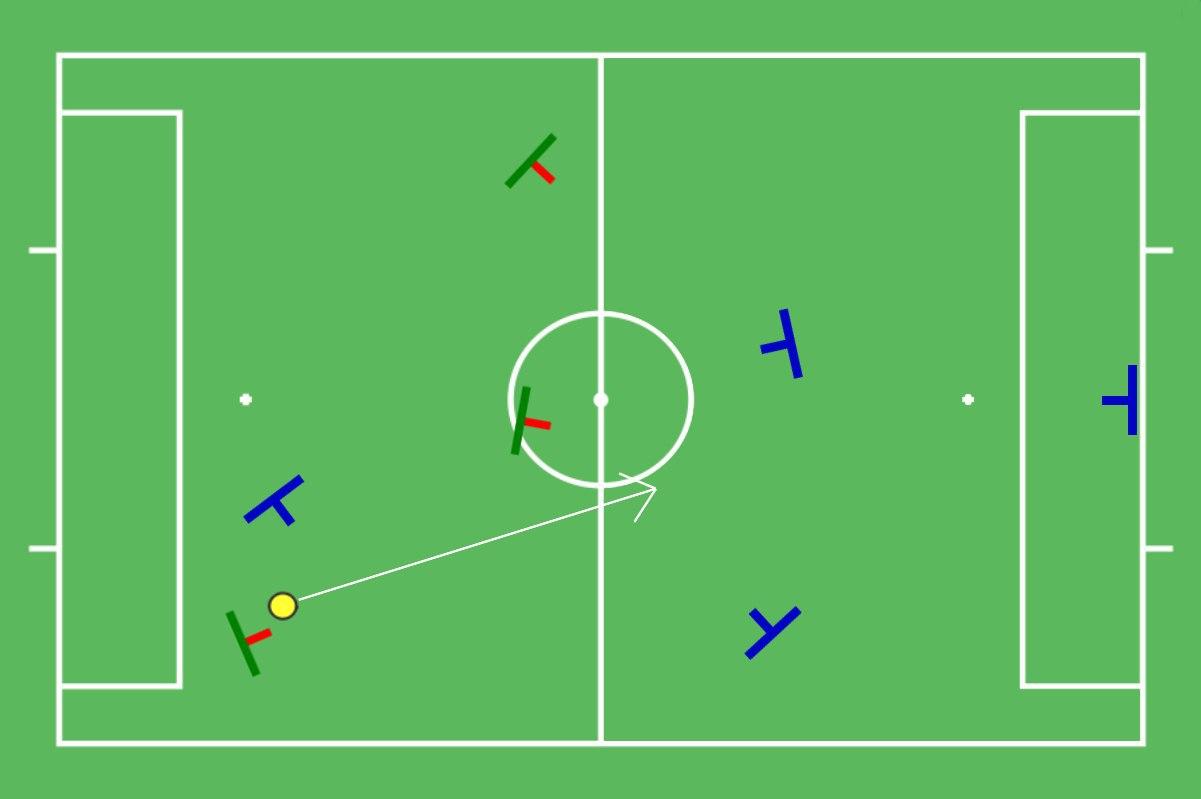}
    \includegraphics[width=\textwidth]{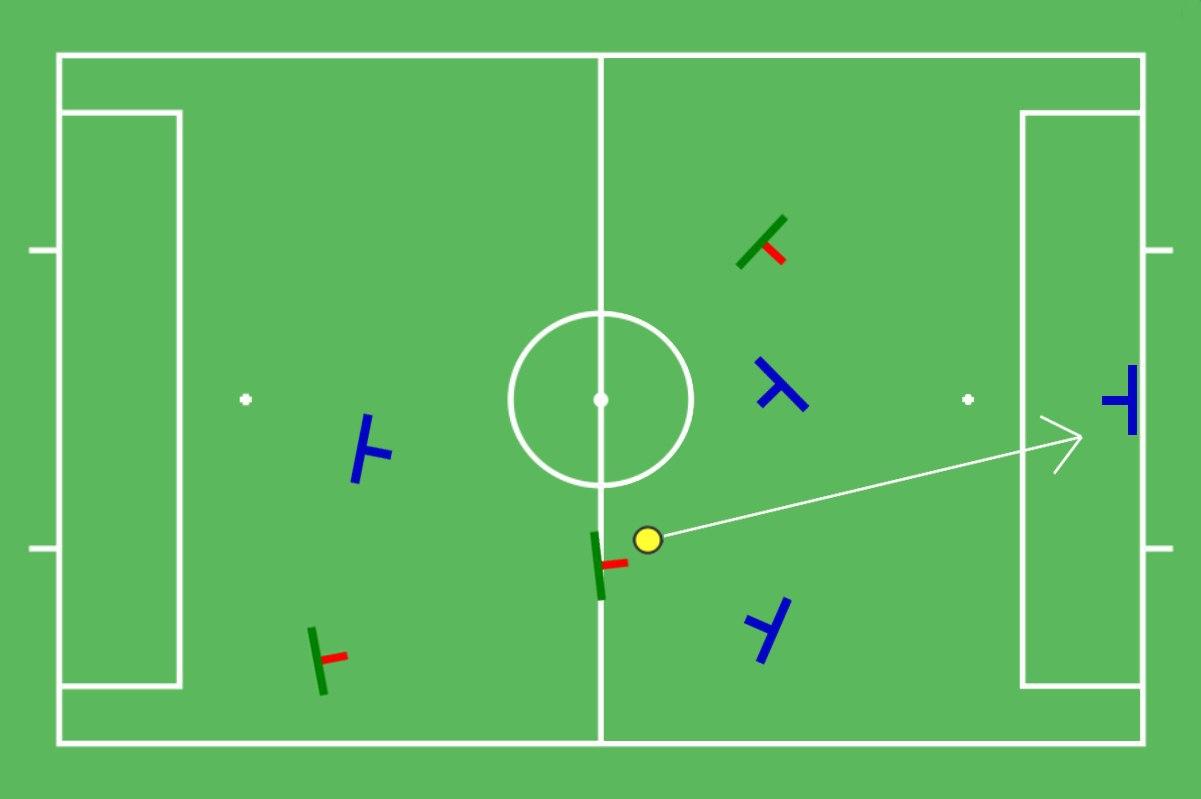}
    \includegraphics[width=\textwidth]{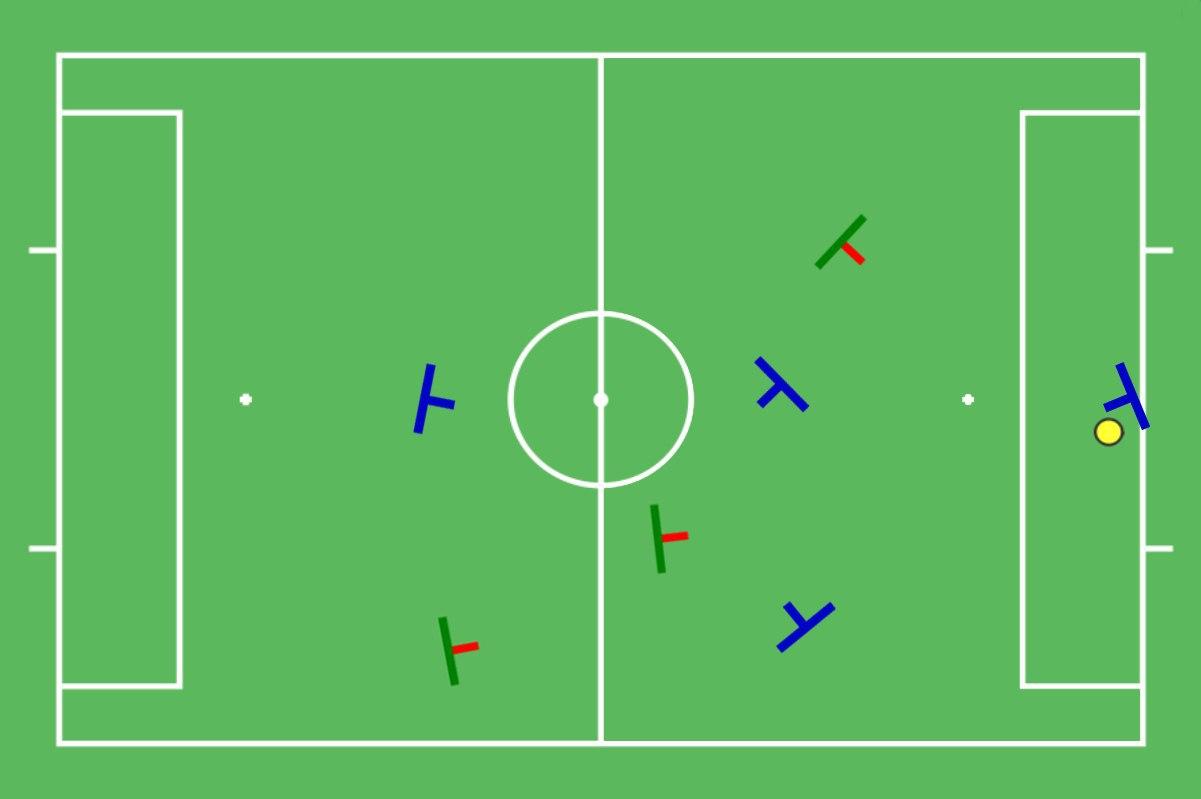}
    \label{fig:expert}
\end{minipage}
\caption{Example of different behaviour resulting from different strategies: planning (left) and expert (right).}
\label{fig:strat}
\end{figure}

Indeed, different strategies lead to different kicks and, a result, to different game outcomes almost always.~\figurename~\ref{fig:strat} shows one such example comparing the most advanced strategies: \textit{expert} and \textit{planning}. Initially both strategies chose nearly the same kick (\figurename~\ref{fig:strat}, top row). However, from then \textit{expert} decided to kick to the goals, while \textit{planning} opted for one extra pass (\figurename~\ref{fig:strat}, middle row). As a result \textit{expert} attack failed -- the final kick was intercepted by a goal-keeper, while \textit{planning} -- did not. In general, qualitative analysis of the recorded games shows that \textit{planning} exploits the pass option much fruitfully than other strategies.

To conduct a quantitative analysis we track the following indicators. First, we tracked the number of games won by a strategy among all the games played -- \textbf{success rate}. Besides, in each game won by our team we measured the following:

\begin{itemize}
    \item \textbf{Time:} Time (simulation time) before goal 
    \item \textbf{Kicks number:} number of kicks in our attack
    \item \textbf{Ball possession, \%:} the ratio of the time that our robot owned the ball to the total time of the attack
\end{itemize}

The last indicator was computed as follows. At each simulated time moment we measured the distance between the ball and the robots. If out robot was close to the ball we considered that our team possessed the ball at that time moment. When the game ended we divided the number of timesteps our team possessed the ball to the total duration of the attack. Additionally in each game (not only won by our team) we measured the number of kicks that passed trough the enemy intersecting zone (recall that with 0.5 probability such a kick was considered to be successfully accomplished) -- \textbf{Intersected}.

\begin{figure}[t]
    \centering
    \includegraphics[width=0.7\textwidth]{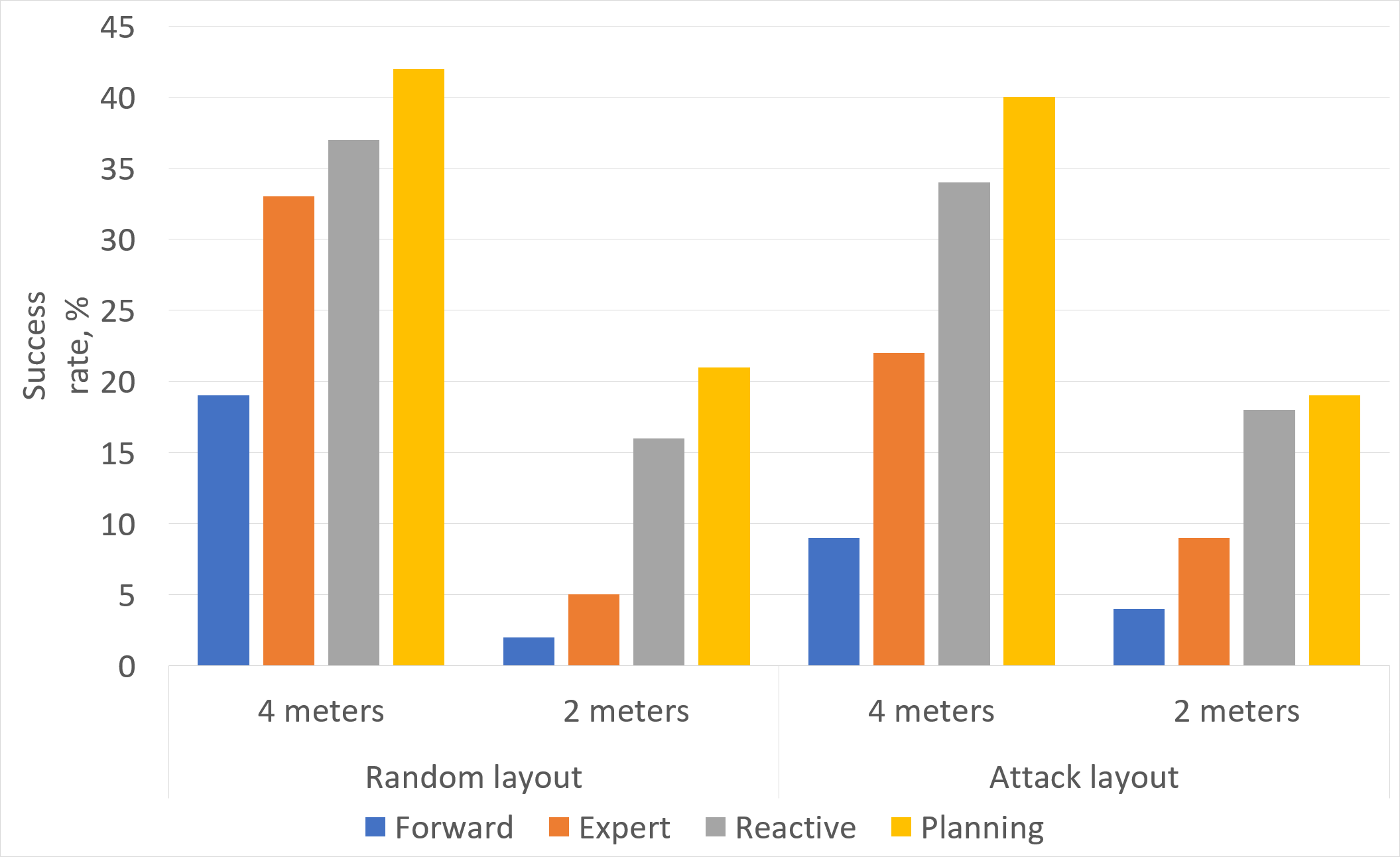}
    \caption{Experiment results: success rate.}
    \label{fig:succ_rate}
\end{figure}

Success rates for different setups are shown in~\figurename~\ref{fig:succ_rate}. As one can \textit{planning} strategy outperformed all the competitors across all the setups. The most pronounceable difference is observed for the \textit{attack} layout with $4~m$ kick length. This is the most important setup from practical point of view. The difference for $2~m$ is less articulated, but ``short'' kicks are not often used in real games.

\tablename~\ref{table:random} provides more statistics for the games when $4~m$ kick was used. Time, kicks number and ball possession indicators were averaged across the games successfully accomplished by expert, reactive and planning strategies. Number of intersected kicks was average across all games played.

For the random layout we see that that team spend on attack slightly decreases for \textit{planning} compared to other strategies. Kicks number is also lower, however the ball possession is better for \textit{expert} strategy. Same holds for the number of intersections. For the attack layout we note the increased number of kicks and, correspondingly, time for the \textit{planning} strategy. This is not an artefact but a quantitative evidence that \textit{planning} utilized idea of pass more often. Observe that the ball possession is nearly the same for all strategies in this layout and in terms of intersections \textit{planning} is the best.

\begin{table}[t]
\centering
\caption{4 meter kick strength statistics}
\begin{tabular}{|c|c|c|c|c||c|c|c|c|}
    \hline
     & \multicolumn{4}{c||}{Random layout} & \multicolumn{4}{|c|}{Attack layout} \\ 
    \hline
    & Forward & Expert & Reactive & Planning & Forward & Expert & Reactive & Planning\\ 
    \hline
    Time, s & - & 11.9 & 11.4 & \textbf{9.8} & - & 27.3 & \textbf{22.9} & 28.7 \\
    \hline
    Kicks number & - & 2.1 & 2.1 & \textbf{2.0} & - & \textbf{3.0} & \textbf{3.0} & 3.5 \\
    \hline
    Ball poss. & - & \textbf{96.85} & 93.77 & 93.94 & - & 99.13 & \textbf{99.49} & 99.26 \\
    \hline
    Intersected & 33 & \textbf{4} & 9 & 9 & 37 & 12 & 10 & \textbf{9}\\
    \hline
\end{tabular}
\label{table:random}
\end{table}

To finalize evaluation we played several full games utilizing RoboCup GameController \footnote{\url{https://github.com/RoboCup-Humanoid-TC/GameController}} according to RoboCup rules. The results again showed the supremacy of the suggested \textit{planning} strategy over the baselines: Game 1: Expert \textbf{1:2} Planning, Game 2: Expert \textbf{0:0} Planning), Game 3: Forward \textbf{0:1} Planning, Game 4: Forward \textbf{0:1} Planning. 

Summarizing the results of the experiments one can claim that the proposed method, indeed, outperforms less advanced approaches across a large variety of game setups. Moreover, qualitative analysis shows that the suggested approach extensively exploits the idea of a pass and can score goals in very complicated layouts, where straightforward approaches do not work.

\FloatBarrier

\section{Conclusions and Future Work}

In this paper we have suggested to utilize heuristic search for planning an attack in humanoid robotic football and introduced all the necessary algorithmic components for that. We evaluated the proposed method in simulation and compared it to the baselines. The former outperformed the latter across a wide range of game scenarios. We plan to integrate the proposed method to the existing game controller software used for MIPT RoboCup team ``Starkit''\footnote{\url{http://starkit.ru}} at the official RocoCup contests. 

An appealing direction of future research is designing a predictive model for robots' behavior and incorporating it to the search algorithm. An orthogonal direction is developing reinforcement learning based planners. Presented work can provide a baseline for the comparison in this case.

\bibliography{bibl.bib}

\end{document}